# Combining chest X-rays and electronic health record (EHR) data using machine learning to diagnose acute respiratory failure

Sarah Jabbour, BSE[1], David Fouhey, PhD[1], Ella Kazerooni, MD[2], Jenna Wiens, PhD[1], Michael W Sjoding, MD[3*]

1. Computer Science and Engineering, University of Michigan, 2260 Hayward St, Ann Arbor, Michigan 48109, USA
2. Department of Radiology, University of Michigan Medical School, 1500 E Medical Center Dr, Ann Arbor, Michigan 48109, USA
3. Pulmonary and Critical Care Medicine, Department of Internal Medicine, University of Michigan Medical School, 1500 E Medical Center Dr, Ann Arbor, Michigan 48109, USA

**ABSTRACT**

**Objective.** When patients develop acute respiratory failure, accurately identifying the underlying etiology is essential for determining the best treatment. However, differentiating between common medical diagnoses can be challenging in clinical practice. Machine learning models could improve medical diagnosis by aiding in the diagnostic evaluation of these patients.

**Materials and Methods.** Machine learning models were trained to predict the common causes of acute respiratory failure (pneumonia, heart failure, and/or COPD). Models were trained using chest radiographs and clinical data from the electronic health record (EHR) and applied to an internal and external cohort.

**Results.** The internal cohort of 1,618 patients included 508 (31%) with pneumonia, 363 (22%) with heart failure, and 137 (8%) with COPD based on physician chart review. A model combining chest radiographs and EHR data outperformed models based on each modality alone. Models had similar or better performance compared to a randomly selected physician reviewer. For pneumonia, the combined model area under the receiver operating characteristic curve (AUROC) was 0.79 (0.77-0.79), image model AUROC was 0.74 (0.72-0.75), and EHR model AUROC was 0.74 (0.70-0.76). For heart failure, combined: 0.83 (0.77-0.84), image: 0.80 (0.71-0.81), and EHR: 0.79 (0.75-0.82). For COPD, combined: AUROC = 0.88 (0.83-0.91), image: 0.83 (0.77-0.89), and EHR: 0.80 (0.76-0.84). In the external cohort, performance was consistent for heart failure and increased for COPD, but declined slightly for pneumonia.

**Conclusions.** Machine learning models combining chest radiographs and EHR data can accurately differentiate between common causes of acute respiratory failure. Further work is needed to determine how these models could act as a diagnostic aid to clinicians in clinical settings.

## INTRODUCTION

Acute respiratory failure develops in over three million patients hospitalized in the US annually.[1] Pneumonia, heart failure, and/or chronic obstructive pulmonary disease (COPD) are three of the most common reasons for acute respiratory failure,[2] and these conditions are among the top reasons for hospitalization for in US.[3] Determining the underlying causes of acute respiratory failure is critically important for guiding treatment decisions, but can be clinically challenging, as initial testing such as brain natriuretic peptide levels or chest radiograph results can be non-specific or difficult to interpret.[4] This is especially true for older adults,[5] patients with comorbid illnesses,[6] or more severe disease.[7] Incorrect initial treatment often occurs, resulting in worse patient outcomes or treatment delays.[8] Artificial intelligence technologies have been proposed as a strategy for improving medical diagnosis by augmenting clinical decision making,[9] and could play a role in the diagnostic evaluation of patients with acute respiratory failure.

Convolutional neural networks (CNNs) are machine learning models that can be trained to identify a wide range of relevant findings on medical images, including chest radiographs.[10] However, for many conditions such as acute respiratory failure, the underlying medical diagnosis is not determined solely based on imaging findings. Patient symptoms, physical exam findings, laboratory results and radiologic results when available are used in combination to determine the underlying cause of acute respiratory failure. Therefore, machine learning models that synthesize chest radiographs findings with clinical data from the electronic health record (EHR) may be best suited to aid clinicians in the diagnosis of these patients. However, efforts to synthesize EHR and imaging data for machine learning applications in healthcare have been limited to date.[11]

We developed a machine learning model combining chest radiographs and clinical data from the EHR to identify pneumonia, heart failure, and COPD in patients hospitalized with acute respiratory failure. We envisioned that such a model could ultimately be used by bedside clinicians as a diagnostic aid in the evaluation of patients with acute respiratory failure, helping them to synthesize multi-modal data and providing estimates of the likelihood of these common conditions. We hypothesized that imaging and clinical data would provide complementary information, resulting in a more accurate model that better replicates the diagnostic process. Finally, we validated the models at an external medical center to determine whether combining these data improves the generalizability of the models.

## METHODS

This study was approved by the Medical School Institutional Review Board at the University of Michigan with a waiver of informed consent among study patients. The study followed the Transparent Reporting of Multivariable Prediction Model for Individual Prognosis or Diagnosis (TRI-POD) reporting guidelines.

### Study population

Models were trained using an internal cohort of patients admitted to an academic medical center in the upper Midwest (Michigan Medicine) in 2017-2018 who developed acute respiratory failure (ARF) during the hospitalization. Models were externally validated on patients admitted to an academic medical center in the northeast (Beth Israel Deaconess Medical Center, BIDMC) in 2014-2016, with clinical data available in the MIMIC-IV dataset[12] and chest radiographs in the MIMIC-CXR dataset.[13-15] In both cohorts, ARF was defined as patients who required significant respiratory support (high flow nasal cannula, noninvasive mechanical ventilation, or invasive mechanical ventilation) and had a chest radiograph performed. We excluded patients who were admitted after routine surgery or a surgical related problem (see Supplement for additional details). The time of ARF diagnosis was defined as when patients first received significant respiratory support.

**Determining the cause of ARF**

To determine the underlying cause of ARF in the Michigan Medicine cohort, physicians independently reviewed the entirety of each patient's hospitalization, including the patient's medical history, physical exam findings, laboratory, echocardiogram, chest imaging results, and response to specific treatments. Patients could be assigned multiple diagnoses if physicians designated multiple causes of ARF, as previous research suggests that multiple concurrent etiologies may be possible.[16] Thus, each physician provided independent ratings of how likely each of the three diagnoses (pneumonia, heart failure, and COPD) was a primary reason for patient's ARF on a scale of 1-4, with 1 being very likely and 4 being unlikely. For patients with multiple reviews, scores were averaged across physicians and patients were assigned the diagnosis if the score was less than 2.5, since 2.5 is the midpoint of 1 and 4. Physician reviewers were board certified in internal medicine (see Supplement **Methods** for further details). We calculated Cohen's kappa[17] and raw agreement rates between physicians due to the difficulty in interpreting of Cohen's kappa in settings of low or high prevalence.[18]

Physician chart review was not performed in the external BIDMC cohort because clinical notes are unavailable in MIMIC-IV. Instead, the cause of ARF was determined based on a combination of International Classification of Disease (ICD)-10 discharge diagnosis codes and medication administration records (Supplement **Table 6-8)**. If a patient had a corresponding ICD-10 code and was treated with medications for a given disease (pneumonia: antibiotics, heart failure: diuretics, COPD: steroids) (Supplement **Table 8**), they were assigned the diagnosis as the etiology of ARF. We also labeled the internal cohort in this manner for a more direct comparison. Accuracy of this approach compared to retrospective chart review was moderate (Supplement **Table 1**).

### Chest radiograph and EHR data extraction and processing

We used chest radiographs nearest to the time of ARF onset (i.e., before or after ARF) in the form of digital imaging and communications in medicine (DICOM) files. Each patient had a corresponding study, containing one or most chest radiographs taken at the same time. Images were preprocessed and downsized to 512x512 pixels, as further described in the Supplement. EHR data included vital signs, laboratory measurements, and demographic data for which a mapping existed between the internal and external cohorts (Supplement **Table 9**). If ARF developed more than 24 hours after admission, we extracted data up until the time of ARF. Because patients frequently present to the hospital with respiratory distress and rapidly develop ARF, clinicians are unable to make a diagnosis until enough data is collected and resulted. Thus, to align with clinical practice, if ARF developed during the first 24 hours of admission, we extracted 24 hours of data to ensure sufficient data for modeling. To avoid temporal information leakage, we excluding variables related to patient treatment, such as medications. Additionally, we selected flowsheet and laboratory data that are commonly performed on all patients with respiratory failure to avoid leaking outcomes. Comorbidity data in the context of diagnosing ARF is typically useful for clinicians when making a diagnosis, but we did not include such data as comorbidities are difficult to capture from EHR data in real-time. In the case of multiple observations for the same variable, the most recent observation to the time of ARF diagnosis was used. Missing data was explicitly encoded as missing, as missingness is likely not at random and has prognostic importance. For example, the presence or absence of a laboratory value (e.g., procalcitonin) might indicate the level of suspicion a physician might have for a particular diagnosis (e.g., pneumonia). We analyze the correlation between missingness and each diagnosis in Supplement **Table 5**. We used FIDDLE, an open-source preprocessing pipeline that transforms structured EHR data into features suitable for machine learning models.[19] After preprocessing, the EHR data were represented by 326 binary features (further described in the Supplement **Methods**).

**Model training**

We trained models to determine the likelihood that pneumonia, heart failure, and/or COPD was an underlying cause of ARF based on clinical data from either the EHR (EHR model), chest radiographs (image model), or both (combined model). The internal cohort was randomly split five times into train (60%), validation (20%), and test (20%) sets. Partitions were made at the patient level such that in each random split, data from the same patient were only in one of the train, validation, and held-out test sets. Separate models were trained on each data split. Additional technical details of model training and architectures are described in Supplement **Figure 1**.

**Model architectures**

EHR model: We trained a logistic regression and two-layer neural network (1 hidden layer, size = 100) with a sigmoid activation to estimate the probability of each diagnosis based on EHR data inputs, treating model type as a hyperparameter. The best EHR model, either logistic regression or two-layer neural network, was chosen based on validation AUROC performance for each of the data splits.

Image model: A CNN with a DenseNet-121[20] architecture was used to estimate the probability of each diagnosis based on the chest radiograph input. The model was first pretrained using chest radiographs from the publicly available CheXpert[10] and MIMIC-CXR-DICOM[21] datasets (excluding patients in the BIDMC validation cohort) to identify common radiographic findings annotated in radiology reports. Then the last layer of the model was fine-tuned to determine ARF diagnoses.[22]

Combined model: Chest radiographs were first passed through the pretrained DenseNet-121 CNN to extract image features. EHR inputs were either passed through a neural network hidden layer or directly concatenated with the extracted image-based features. The presence or absence of the EHR input hidden layer prior to concatenation was treated as a model hyperparameter. Finally, the concatenation was passed through an output layer with a sigmoid activation to estimate the probability of each diagnosis. Like the image model, parameters of the DenseNet-121 were frozen after pretraining.

**Model evaluation**

We evaluated the value of combining chest radiographs and EHR data by comparing the combined model to the EHR and image models in terms of the individual and macro-average AUROC for pneumonia, heart failure, and COPD when applied to the internal MM cohort test sets. The median and range of model performance on the internal cohort test sets are reported across the five splits. We calculated the area under the precision recall curve (AUPR) in a similar manner. We also measured calibration performance by calculating the expected calibration error (ECE) and generated calibration plots.[23] We also calculated diagnostic test metrics for each model including sensitivity, specificity, and the diagnostic odds ratio at a positive predictive value (PPV) of 0.5 for each condition.

We compared model performance to that of a randomly selected physician reviewer on patients that underwent three or more physician chart reviews. This evaluation required that we change the “ground truth” label so the randomly selected physician reviewer was not used to generate the ground truth label. To calculate physician performance, we compared a randomly selected physician to all the other physicians who reviewed the same patient. The new “ground truth” label was then calculated as the average of the remaining reviews for each patient. The combined model was then compared to a randomly selected physician in terms of individual and macro-average AUROC.

To understand the generalizability of the models, we applied each of the five models trained on MM to the external BIDMC cohort, calculating performance based in terms of the individual and macro-average AUROC on diagnosis codes. To compare performance across cohorts, we compared results based on ICD-10 codes and medications in both cohorts.

**Feature Importance**

Since large capacity models are known to pick up on spurious features,[24] we performed a feature importance analysis to understand how our models used chest radiograph and EHR data to make predictions. For chest radiographs, heatmaps were generated to understand which regions of the chest radiograph influenced the model prediction.[25] To highlight the most important regions in each image, heatmaps were normalized on a per-image basis. We qualitatively reviewed all heatmaps and identified high level patterns. Randomly selected patients are shown for illustrative purposes from the group where both the image and combined models either correctly classified or incorrectly classified the diagnosis and were most confident in their predictions (i.e., those patients whose predictions were in the top 85$^{th}$ percentile of predictions in the internal cohort test set).

To understand which EHR features were important in model decisions, we measured permutation importance. We grouped highly correlated variables together (Pearson’s correlation > 0.6). Features were ranked from most to least important based on the drop in AUROC when these features were randomly shuffled across examples in the test set.[26] We averaged feature rankings across all five test sets and report the five highest ranked features for each diagnosis.

## RESULTS

### Study population

The internal cohort included 1,618 patients, with 666 (41%) females and a median age of 63 years (IQR: 52-72). The external cohort demographics were similar, although there was a higher percentage of patients in the other or unknown race categories (**Table 1**). In the internal cohort, 29% of patients were reviewed by three or more physicians, 48% by 2 physicians, and 23% of patients were received by 1 physician. Based on chart review, there were 508 (31%) patients with pneumonia, 363 (22%) with heart failure and 137 (9%) with COPD as the underlying cause of ARF. More than one of these diagnoses were present in 155 (10%) patients. Raw agreement between reviewers was 0.78, 0.79, and 0.94 for pneumonia, heart failure, and COPD respectively and Cohen's kappa was 0.47, 0.48, and 0.56 respectively, which is slightly higher than previous publications (Supplement **Table 2**).[27-29] The prevalence of pneumonia, heart failure, and COPD was lower in the external cohort compared to the internal cohort when diagnoses were determined based solely on diagnosis codes and medication administration records (**Table 1**).

**Table 1.** Characteristics of the internal and external cohorts.

| Characteristic | Internal cohort (n=1618) | External cohort (n=1774) |
|---|---|---|
| Age, median (IQR) | 63 (52-72) | 63 (48-75) |
| Gender, n (%) | | |
| Male | 952 (59) | 1020 (57) |
| Female | 666 (41) | 754 (43) |
| Race, n (%) | | |
| White | 1364 (84) | 904 (51) |
| Black | 159 (10) | 151 (9) |
| Other/Unknown | 95 (6) | 719 (41) |
| Acute Respiratory Failure Etiology, n (%) | | |
| Pneumonia | 508 (31) | NA |
| Heart Failure | 363 (22) | NA |
| COPD | 137 (9) | NA |
| Pneumonia & Heart Failure | 82 (5) | NA |
| Pneumonia & COPD | 64 (4) | NA |
| COPD & Heart Failure | 35 (2) | NA |
| All Conditions | 13 (1) | |
| Diagnosis codes + Medications, n (%) | | |
| Pneumonia | 650 (40) | 322 (18) |
| Heart Failure | 413 (26) | 204 (11) |
| COPD | 244 (15) | 70 (4) |
| Pneumonia & Heart Failure | 185 (11) | 103 (6) |
| Pneumonia & COPD | 127 (8) | 46 (3) |
| COPD & Heart Failure | 106 (7) | 29 (2) |
| All Conditions | 56 (3) | 21 (1) |

Acute respiratory failure etiology was determined based on retrospective chart review performed by one or more physicians. Diagnosis codes are the International Classification of Disease-10 diagnosis codes assigned to the hospitalization. Abbreviations: NA: not available; IQR: interquartile range; COPD: chronic obstructive pulmonary disease.

**Model performance on the internal cohort**

The combined model demonstrated a higher macro-average AUROC (AUROC = 0.82, range: 0.80-0.85) compared to the image model (AUROC = 0.78, range: 0.75-0.81) and EHR models (AUROC = 0.77, range: 0.76-0.80) (**Figure 1, Table 3**). The combined model was more accurate than the image and EHR models for all three diagnoses. The combined model also had a higher macro-average area under the precision recall curve (AUPR), (AUPR = 0.64, range: 0.55-0.67) compared to the image (AUPR = 0.53, range: 0.46-0.57) and EHR models (AUPR = 0.51, range: 0.48-0.53), and a higher AUPR for all individual diagnoses (Supplement **Table 3**). All models demonstrated fair calibration as measured by the expected calibration error (Supplement **Figure 3**).

The combined model outperformed the image and EHR models in terms of sensitivity and diagnostic odds ratio (**Table 2**). The combined model's diagnostic odds ratio was 5.79 (range: 4.90-6.42) for pneumonia, 7.85 (range: 5.61-10.20) for heart failure, and 37.00 (20.50-54.80) for COPD (**Table 2**).

**Table 2.** Sensitivity, specificity, and diagnostic odds ratio of all models in the internal cohort.

| Model and diagnosis | Sensitivity % (range) | Specificity % (range) | Diagnostic odds ratio (range) |
|---|---|---|---|
| **Combined** | | | |
| Pneumonia | 81 (71-85) | 60 (50-70) | 5.79 (4.90-6.42) |
| Heart failure | 62 (53-71) | 83 (76-85) | 7.85 (5.61-10.20) |
| COPD | 68 (44-81) | 94 (93-97) | 37.00 (20.50-54.80) |
| **Image** | | | |
| Pneumonia | 65 (60-78) | 69 (55-75) | 4.28 (3.96-4.64) |
| Heart failure | 52 (41-67) | 86 (80-87) | 6.71 (4.65-9.42) |
| COPD | 41 (12-54) | 97 (95-99) | 21.60 (13.40-29.80) |
| **EHR** | | | |
| Pneumonia | 64 (63-84) | 69 (56-73) | 4.04 (3.42-6.83) |
| Heart failure | 56 (45-67) | 85 (79-88) | 7.35 (5.91-7.74) |
| COPD | 29 (4-48) | 98 (96-100) | 18.00 (10.90-25.00) |

Sensitivity, specificity, and diagnostic odd ratio are calculated at a PPV of 0.5 for the internal cohort based on physician chart review. Abbreviations: COPD: chronic obstructive pulmonary disease.

**Table 3**. Performance of image, EHR and combined models on the internal held-out test set and external validation cohort in terms of AUROC.

| Cohort and Model | Pneumonia | Heart Failure | COPD | Macro-Average AUROC |
|---|---|---|---|---|
| **Internal chart review (n, % pos)** | **324 (322-324) 32% (29-36)** | **324 (322-324) 21% (20-24)** | **324 (322-324) 8% (5-)** | **--** |
| Image | 0.74 (0.72-0.75) | 0.80 (0.71-0.81) | 0.83 (0.77-0.89) | 0.78 (0.75-0.81) |
| EHR | 0.74 (0.70-0.76) | 0.79 (0.75-0.82) | 0.80 (0.76-0.84) | 0.77 (0.76-0.80) |
| Combined | 0.79 (0.77-0.79) | 0.83 (0.77-0.84) | 0.88 (0.83-0.91) | 0.82 (0.80-0.85) |
| **Internal diagnosis codes + meds (n, % pos)** | **324 (322-324) 45% (37-46)** | **324 (322-324) 26% (22-29)** | **324 (322-324) 15% (15-16)** | **--** |
| Image | 0.67 (0.62-0.71) | 0.79 (0.78-0.80) | 0.69 (0.68-0.71) | 0.72 (0.70-0.73) |
| EHR | 0.66 (0.64-0.74) | 0.78 (0.77-0.83) | 0.72 (0.70-0.80) | 0.74 (0.71-0.76) |
| Combined | 0.71 (0.65-0.75) | 0.82 (0.81-0.83) | 0.76 (0.73-0.79) | 0.76 (0.75-0.78) |
| **External diagnosis codes + meds (n, % pos)** | **n=1774 18%** | **n=1774 11%** | **n=1774 4%** | **--** |
| Image | 0.64 (0.63-0.65) | 0.81 (0.80-0.82) | 0.81 (0.78-0.82) | 0.75 (0.75-0.76) |
| EHR | 0.62 (0.60-0.63) | 0.76 (0.70-0.77) | 0.78 (0.74-0.79) | 0.72 (0.69-0.73) |
| Combined | 0.65 (0.64-0.66) | 0.82 (0.81-0.84) | 0.86 (0.86-0.86) | 0.78 (0.77-0.78) |

Performance as determined based on the AUROC. The internal cohort was randomly split five times into train (60%), validation (20%) and test (20%) sets. The median AUROC and AUROC range are reported for models trained on each split. The resulting five models were applied to the external cohort and the median AUROC and AUROC range are reported for models. Abbreviations: AUROC: area under the receiver operating characteristic; COPD: chronic obstructive pulmonary disease.

**Model performance compared to a randomly selected physician**

The combined model demonstrated similar or better performance in terms of individual and macro-average AUROC for all three diagnoses compared to randomly selected physicians (**Table 4**). For pneumonia, the combined model AUROC = 0.74 (range: 0.68-0.84) and the physician AUROC = 0.75 (range: 0.73-0.83); for heart failure, the combined model AUROC = 0.79 (range: 0.75-0.87) and physician AUROC = 0.77 (range: 0.73-0.84): for COPD, the combined model AUROC = 0.89 (range: 0.71-0.98) and physician AUROC = 0.78 (range: 0.72-0.88).

**Table 4**. Comparison of the combined model to a randomly selected physician

| Cohort and Model | Pneumonia | Heart Failure | COPD | Macro-Average AUROC |
|---|---|---|---|---|
| **(n, % pos)** | **98 (90-100)<br>36% (33-42)** | **98 (90-100)<br>26% (28-38)** | **98 (90-100)<br>6% (4-11)** | **--** |
| Randomly selected physician | 0.75 (0.73-0.83) | 0.77 (0.73-0.84) | 0.78 (0.72-0.88) | 0.79 (0.75-0.81) |
| Combined model | 0.74 (0.68-0.84) | 0.79 (0.75-0.87) | 0.89 (0.71-0.98) | 0.84 (0.76-0.85) |

Analysis performed in patients with three or more physician chart reviews. For each patient, one physician reviewer was randomly selected and compared to the model. The “ground truth” label used was calculated as the average of the remaining reviewers for each patient. Median performance and ranges are reported across 5 data splits.
Abbreviations: COPD: chronic obstructive pulmonary disease. AUROC: Area under the receiver operator characteristic curve.

**Model performance in the external cohort**

The combined model was consistently more accurate than other models in terms of AUROC (**Figure 1**, **Table 3**). The image model consistently outperformed the EHR model for all three diagnoses. When comparing the performance of the model across centers using diagnosis codes and medication administration as the “gold standard,” there was no change in the combined model AUROC performance for heart failure (median AUROC = 0.82), and an increase in performance for COPD (median AUROC increasing from 0.76 to 0.86 for COPD), suggesting transferability. However, the decline for pneumonia was more substantial (0.71 to 0.65).

**Understanding model decisions**

Both the image and combined models focused on appropriate areas on the chest radiograph when correctly diagnosing heart failure and pneumonia, including the lungs and heart (**Figure 2**). When correctly diagnosing COPD, models appeared to focus on the trachea. In cases where models made incorrect diagnoses, they still focused on appropriate anatomical areas (Supplement **Figure 2**).

The EHR and combined models were influenced by similar clinical features with some deviations (**Table 5**). In most cases, important clinical features identified by the model aligned with the clinical understanding of diagnosis. For pneumonia, the oxygen saturation, procalcitonin level and troponin were important variables. For heart failure, brain natriuretic peptide (BNP), troponin and patient age were important variables. In contrast, variables identified as important for identifying COPD were less closely aligned with clinical understanding of diagnosis for COPD, such as mean corpuscular hemoglobin (MCHC) or magnesium.

**Table 5.** Top five important clinical features used by the EHR and combined models to identify etiologies of acute respiratory failure.

| Diagnosis | EHR model | Combined model |
|---|---|---|
| Pneumonia | Oxygen saturation or $PaO_2$ | Oxygen saturation or $PaO_2$ |
| | Procalcitonin | Procalcitonin |
| | Troponin-I | Troponin-I |
| | Absolute lymphocyte count | Plateau pressure* |
| | Plateau pressure* | BNP |
| Heart Failure | BUN or Creatinine | BUN or Creatinine |
| | BNP | Troponin-I |
| | Troponin-I | BNP |
| | Tidal Volume* | Tidal Volume* |
| | Age | Age |
| COPD | MCHC | MCHC |
| | Oxygen saturation or $PaO_2$ | Total bilirubin |
| | Lymphocytes % or Neutrophils % | Bicarbonate |
| | Bicarbonate | Magnesium |
| | Age | Alkaline phosphate |

Top features identified by permutation importance. Highly correlated features (>0.6) were grouped together during the permutation importance analysis and reported together (e.g. BUN or Creatinine). *Plateau pressure and tidal volume measured during invasive mechanical ventilation. Abbreviations: $PaO_2$: Partial pressure of oxygen; MCHC: mean corpuscular hemoglobin concentration; BUN: Blood urea nitrogen; BNP: Brain natriuretic peptide.

## DISCUSSION

We developed and validated machine learning models combining chest radiographs and clinical data to determine the underlying etiology of patients with ARF. Overall, the models combining chest radiographs and clinical data had better discriminative performance on both internal and external validation cohorts compared to models analyzing each data type alone. They also demonstrated similar or better performance compared to randomly selected physician reviewers. Physician reviewers had access to substantially more information than the model, including patient history, physical exam findings, and response to specific treatments. Thus, its notable that the model can match physician performance. Given the diagnostic challenges of determining the underlying etiology of ARF in practice, such models have the potential to aid clinicians in their diagnosis of these patients.

Many studies of machine learning applied to chest radiographs have used a radiologist interpretation of chest radiology studies to train models.[10] However, for medical conditions including pneumonia, heart failure, or COPD, a clinical diagnosis is not determined solely based on chest radiographic findings. The underlying diagnosis is based on a combination of concordant clinical symptoms (e.g., productive cough), physical examination findings, laboratory results, and radiologic imaging findings when available. Our models more closely resemble clinical practice, since they combine chest radiographs and other clinical data, and were also trained using diagnoses determined by physicians who reviewed the entirety of each patient's hospitalization, rather than just chest radiographs alone.

Improving clinical diagnosis has been identified as important for improving healthcare quality,[30] and machine learning could support the diagnostic process in several ways. First, clinicians may over focus on certain clinical data (e.g., BNP value when diagnosing heart failure) or may be prone to other cognitive errors.[31] Models may provide more consistent estimates of disease probabilities based on the same data (though may be prone to other errors as discussed below). Second, models may identify features not typically considered by clinicians. For example, when diagnosing COPD, our models frequently focused on the tracheal region, whereas clinical references do not emphasize radiology findings.[32] Yet, tracheal narrowing (i.e., "saber-sheath" trachea) can be a marker of severe air-flow obstruction,[33] so training clinicians to look for this feature might also be useful. Radiologists may only apply criteria for reporting a saber-sheath trachea in severe cases, with milder transverse narrowing on front chest radiographs not considered specific enough for a diagnosis of COPD.

Importantly, the machine learning models presented in this paper are not envisioned to replace clinicians, but rather to serve as a *diagnostic aid*: providing additional information similar to diagnostic tests which could result in quicker diagnosis and treatment. Clinicians have access to important diagnostic data such as subjective patient complaints or physical exam findings that are not readily available as model inputs. Thus, collaborations between clinicians and models, where clinicians consider model results in the full context of the patient's hospitalization, could be an optimal use of such models. One caveat is that models may also use shortcuts,[24] i.e., take advantage of spurious correlations in the training data that might not hold across populations. Clinicians might be able to recognize when a model is taking a shortcut and discount the model's output in such settings. For example, we noted that our model focused on the presence of pacemakers for heart failure (similar to Seah *et al.*[34]), which may lead it to perform poorly in heart failure subpopulations without pacemakers, or to overestimate the probability of heart failure when other data would suggest an alternative diagnosis. Similarly, since there are no established EHR markers for COPD, the clinical variables the model identified as important in COPD might not align with clinical intuition and could be noise in the data. Further investigation of these identified COPD features is needed for confirmation.

However, there are still several scenarios where this model may provide clinical benefit. While the model identifies many features that are already well-known to bedside clinicians for diagnosis, it is also capable of synthesizing many more features than a clinician can. Thus, it may be useful in straightforward diagnostic cases where a clinician might be busy, distracted, or unable to effectively synthesize the entirety of all available information at once. Additionally, the model may also improve diagnostic accuracy in difficult cases. Clinicians may make diagnostic and treatment errors in up to 30% of patients.[35] The combined model exhibits similar or improved performance compared to a randomly selected physician for all three diagnoses. Nonetheless, such a model would need to be carefully integrated into clinical workflows to support the diagnostic process. Studying the implementation of models combining chest radiographs and EHR data is important and necessary future work.

Our study has limitations. We used a limited set of EHR inputs that are commonly collected in all patients with respiratory failure, easily transferred across institutions, and excluded variables related to patient treatment decisions limit the model's ability to learn the underlying diagnosis of ARF by learning clinician actions. However, we are unable to fully exclude the possibility that some variables used could indirectly allude to patient treatment. We also designed the model to use EHR data that is readily available in real-time. Since analysis of comorbidity data is most often based on hospital diagnosis codes which are generated after the hospitalization and are inconsistent across institutions, we did not include comorbidity in the model.[36,37] Knowledge of comorbidities (e.g. prior history of COPD) is useful to for diagnosis, therefore, future efforts to make comorbidity data available to models when running in real-time is warranted.

We made other modeling choices and make our code available so others can investigate alternative approaches. We also used a simple architecture that concatenated EHR and image features, which may prevent the network from using the EHR data as guidance when extracting features from the chest radiographs earlier in the network. However, introducing EHR data at the beginning of the network requires retraining the large DenseNet-121 network, which is likely infeasible given the limited training data in the current study. While pretraining was used to enhance model performance, this does not rule out the possibility of negative transfer.[38] More pretraining data specific to the diagnostic task could improve performance as well as model pretraining that includes both structured clinical and imaging data. We also ignore the temporal ordering of the EHR data (i.e., using only the most recent, rather than all measurements), which may miss some relevant diagnostic information or trends.

Finally, we used two different methods of determining patient diagnoses to evaluate model performance. First, chart reviews performed by multiple physicians was used to determine the ground truth diagnosis of patients in the internal cohort. While such labelling is imperfect, multiple reviews were averaged when available to improve diagnostic accuracy. In this way, the model can be thought of as being trained to learn the collective expertise of multiple physicians. Second, because diagnosis codes may only be moderately aligned with the actual clinical diagnosis,[37] we used both diagnosis codes and medications which may be a better proxy for diagnoses in the external cohort since we did not have access to clinical notes to conduct chart review. Despite the potential for differences in diagnosis labels across institutions, model performance did not drop for heart failure and COPD. Ultimately, prospective model validation will be needed to determine the model's performance in practice and its ability to support clinicians in the diagnostic process.

In summary, machine learning models leveraging both chest radiographs and EHR data can accurately differentiate between common causes of ARF (pneumonia, heart failure, and/or COPD) and generalize better to another institution compared to models using only radiographic or EHR data alone. These findings highlight the potential of machine learning to aid in the clinical diagnoses of pneumonia, heart failure, and COPD. Combined with the expertise of clinicians, such models could improve the diagnostic accuracy of clinicians in this challenging clinical problem.

**FUNDING**

This work was supported in part by grants from the National Institutes of Health (K01HL136687, R01 HL158626, and R01 LM013325) and a University of Michigan Precision Health Award.

**CONTRIBUTORS**

Concept and design: all authors

Acquisition, analysis, or interpretation of data: all authors

Drafting of the manuscript: Jabbour

Critical revision of the manuscript for important intellectual content: all authors

Statistical analysis: all authors

Obtained funding: Sjoding, Wiens, Fouhey

Study supervision: Sjoding

**CONFLICT OF INTEREST**

All authors report no competing interests to disclose.

**ACKNOWLEDGEMENTS**

This work was supported in part by grants from the National Institutes of Health (K01HL136687, R01 HL158626, and R01 LM013325) and a University of Michigan Precision Health Award.

**DATA SHARING**

The code for this study, along with the trained weights of the resulting models, are available at https://github.com/MLD3/Combining-chest-X-rays-and-EHR-data-ARF. The CheXpert dataset used in this study can be accessed from https://stanfordmlgroup.github.io/competitions/chexpert/. The MIMIC-IV chest x-ray and clinical dataset can be accessed at https://physionet.org/content/mimiciv/1.0/ and https://physionet.org/content/mimic-cxr/2.0.0/. Data from the University of Michigan are not publicly available. A limited, de-identified version could be made available to other researchers from accredited research institutions after entering into a data use agreement with the University of Michigan.

## FIGURES

**Figure 1. Performance of the combined, image, EHR model for diagnosis of pneumonia, heart failure and COPD in the internal and external cohorts.**

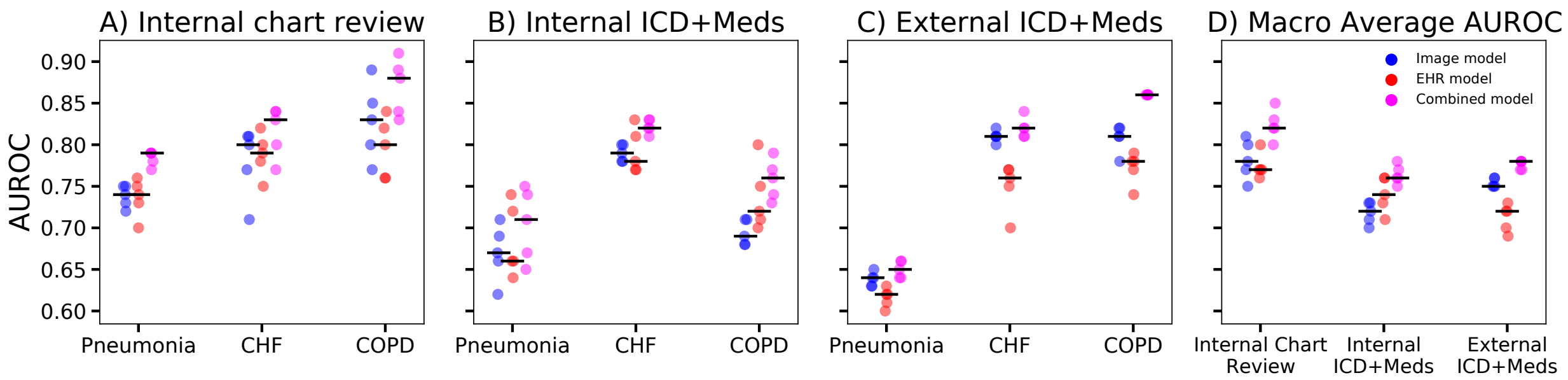

Model performance evaluated based on the area under the receiver operator characteristic curve (AUROC). Black horizontal lines indicate median performance for each model. When the models were evaluated using diagnosis based on chart reviews in the internal cohort, the combined model outperforms the image and EHR models on most data splits in terms of AUROC for identifying pneumonia and COPD, and better for one of the five data splits for diagnosing heart failure (a). Model performance decreased for pneumonia and COPD when evaluated using on discharge diagnosis codes and medications (b). Model performance on the external cohort was evaluated using discharge diagnosis codes and medications (c) and was similar to the internal cohort (b) with the exception of pneumonia. The combined model consistently outperformed the other models across cohorts in terms of macro-average AUROC which combines model performance across all three diagnoses. Abbreviations: COPD: chronic obstructive pulmonary disease.

**Figure 2. Chest radiograph heatmaps in patients where the model correctly diagnosed pneumonia, heart failure or COPD with high probability.**

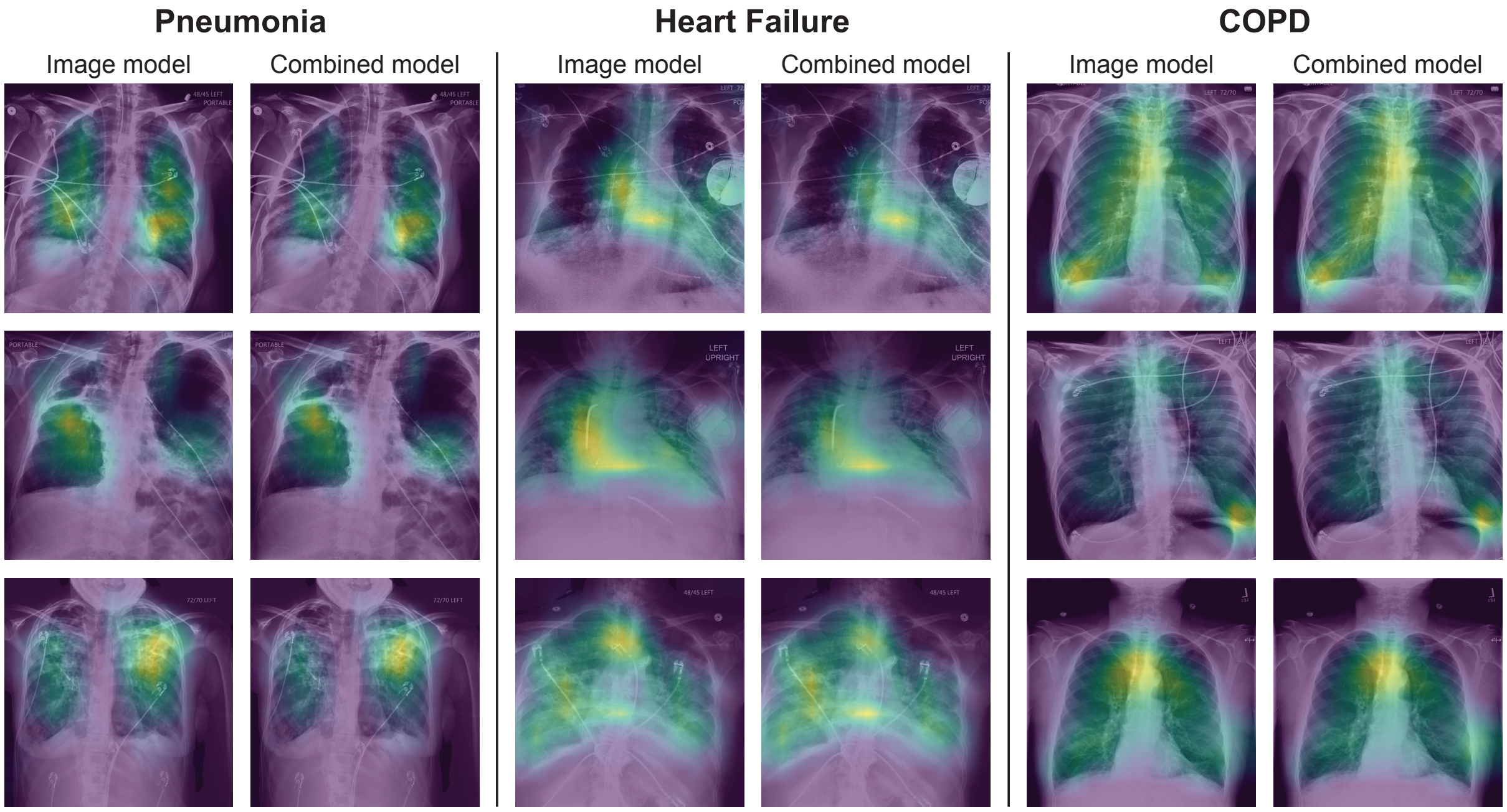

Chest radiographs are shown for patients that the model correctly classified as positive for each disease with high probability. The overlaying heatmap generated by Grad-CAM highlights the regions the model focused on when estimating the likely diagnosis (blue: low contribution, yellow: high contribution). For both the image and combined model, the model looked at the lung regions for pneumonia and COPD and the heart region for heart failure. Heatmaps were normalized on individual images to highlight the most important areas of each image, therefore heatmap values should not be compared across images. Image processing was performed, including histogram equalization to increase contrast in the original images, and then images were resized to 512x512 pixels.

## SUPPLEMENT

## Combining chest X-rays and EHR data using machine learning to diagnose acute respiratory failure

## Methods

### Physician Reviewers

There were 17 physicians board certified in internal medicine who performed reviews, including 6 pulmonary and critical care faculty, 10 pulmonary and critical care fellows, and 1 hospitalist. All pulmonary fellows have previously completed a three-year internal medicine residency and at least one year of pulmonary and critical care fellowship. 49% of the 3,741 total reviewers were performed by pulmonary and critical care fellows, 40% were performed by pulmonary and critical care faculty, and 11% were performed by a practicing hospitalist who was board certified in internal medicine. Clinical reviewers were unaware of the model predictions at the time they rated the likelihood of each diagnosis.

### Cohort selection

Michigan Medicine (MM) internal cohort: Patients were included if they developed acute respiratory failure during hospitalizations in 2016 and 2017. Acute respiratory failure was defined as the need for high flow nasal cannula, endotracheal tube, or bipap mask based on respiratory flowsheet documentation during the first 7 days of their hospitalization. Patients were excluded if they were admitted to the neurologic or cardiovascular vascular ICU after a surgical procedure. A formal sample size calculation was not performed.

Beth Israel Deaconess Medical Center (BIDMC) external cohort: Patients were included if they developed acute respiratory failure during hospitalization in 2014-2016. Specifically, they received supplemental oxygen in the form of high flow nasal cannula, endotracheal tube, or bipap mask and had linked chest radiographic images in the MIMIC-CXR dataset and clinical data in the MIMIC-IV dataset. Patients who were admitted for a surgical related problem were excluded. More specifically, patients were excluded if they received oxygen support while admitted under a surgical unit (CSURG, NSURG, ORTHO, SURG, TSURG, or VSURG) or within 24 hours after leaving a surgical unit. Like the MM cohort, a formal sample size calculation was not performed.

### Additional details of data extraction and preprocessing

Chest radiographs and EHR data were automatically extracted without any clinical assessment of the two predictors.

After obtaining chest radiographic images in the form of digital imaging and communications in medicine (DICOM) files, global histogram equalization was first applied to the images to increase contrast in the original images. Then, images were resized, while preserving their aspect ratio, such that their smaller axis was 512 pixels. We randomly cropped the training images to 512x512 and augmented with random in-plane rotations up to 15 degrees. Validation and test images were center cropped to 512x512.

To process the EHR data, we used FIDDLE, an open-source preprocessing pipeline that transforms structured EHR data into feature vectors suitable for machine learning models.[1] FIDDLE maps all variables (e.g., temperature = 37℃) into five binary features (e.g., [0,1,0,0,0]) corresponding to ranges of values (e.g., [35-36, 37-38, 39-40, 40-41, 41-42]), and accounts for missingness by setting all values in the feature vector to zero. After preprocessing, the EHR data were represented by 326 binary features.

### Model training details and hyperparameter tuning

In all cases, model parameters were learned using stochastic gradient descent with momentum to minimize cross-entropy loss based on the chart review diagnostic labels. Final models and hyperparameters including the learning rate, momentum, and weight decay were selected based on validation AUROC performance. To prevent overfitting, we applied L2-regularization to all learned model parameters, and early stopping was used with a patience of 5 epochs. Additionally, for both the image and combined models, parameters of the DenseNet-121 were frozen after pretraining. Finally, we compared final model performance in the train, validation, and test sets to further assess for overfitting (Supplement **Table 4**). For both the EHR and combined models, a drop in performance between the train and validation/test sets was observed, while minimal changes in performance were observed for the Image model. This suggests some overfitting to the clinical data. Additionally, since some patients had multiple chest radiographs taken at the same time, models were applied to all chest radiographs and the predictions were averaged. We swept the learning rate from [$10^{-4}$, $10^{-3}$, $10^{-2}$, $10^{-1}$,1, 3], momentum from 0.8 and 0.9, and weight decay from [$10^{-4}$, $10^{-3}$, $10^{-2}$, $10^{-1}$]. A batch size of 32 was used throughout. For the EHR model, two architectures were swept: one and two-layer (hidden units: 100, ReLu activation) neural networks with sigmoid output activation. Similarly for the combined model, two architectures were swept: one where EHR data were concatenated with Image-based features, and one with a hidden layer (hidden units: 100, ReLu activation) after EHR data before concatenation with Image-based features.

**Model initialization**

We initialized the DenseNet-121 model first using pretrained weights on CheXpert[2] and then MIMIC-CXR[3] chest radiographs that were excluded from the external validation cohort. Histogram equalization was first applied to the images to increase contrast in the original images, and then images were resized such that their smaller axis was $512$ pixels while preserving their aspect ratio. This allowed for cropping the images horizontally or vertically to a square 512x512 image as input to the model. We trained the model to predict text-mined radiology report labels and optimized the sum of the *masked* binary cross-entropy loss across labels, masking the loss for labels with a missing value. Following Irvin *et al.,* we used Adam with default parameters of $\beta_1 = 0.9$ and $\beta_2 = 0.999$, learning rate of $10^{-4}$, and a batch size of 16.[2] We trained for 3 epochs with 3 different random initializations, saving checkpoints every 4,800 batches. We first trained on the CheXpert data and selected the checkpoint that performed the best on a the CheXpert validation set of size 200, measured by average area under the receiver operator characteristic curve (AUROC) across all 14 labels. Then, we trained on the MIMIC-CXR data and again selected the checkpoint that performed the best on a randomly sampled validation set of size 5000.

**Model Calibration**

We calibrated the image, EHR, and combined models based on the internal MM validation set. For each model and each disease, we divided the model predictions into quintiles and calculated the average prediction of each quintile. We then plotted the average prediction over the true probability of disease and computed a line of best fit to map the models' predictions to calibrated predictions. Finally, we measured the expected calibration error (ECE) based on quintiles of predicted risk by calculating the average absolute difference between predicted risk and observed risk (Supplement **Figure 3)**.[4]

## Model Architectures

**Figure 1**. Model architectures.

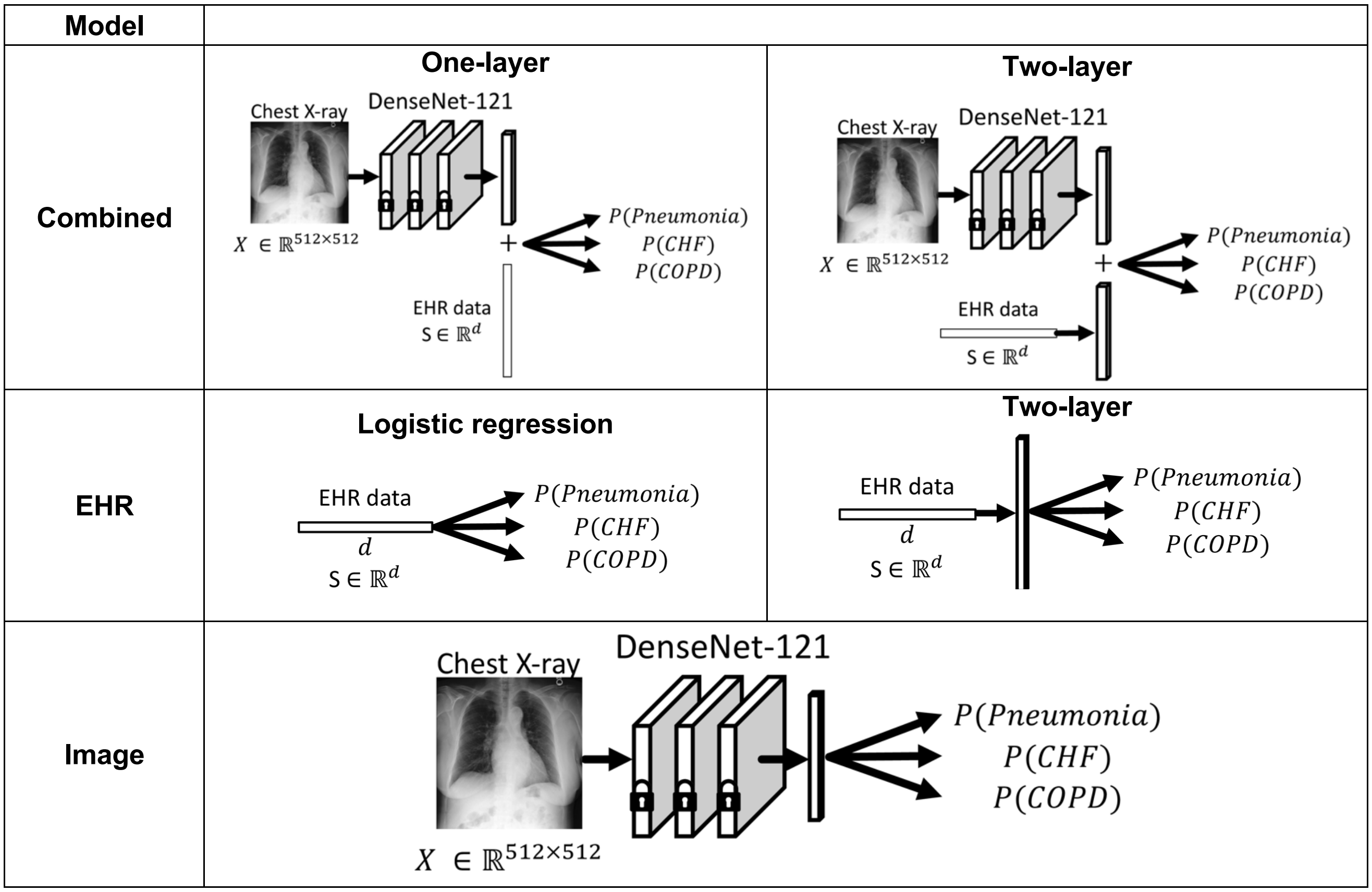

The image model consisted of a DenseNet-121 with a final fully connected layer and sigmoid output activation for each diagnosis. The EHR model consists of a one-layer or two-layer neural network with sigmoid output activation for each diagnosis. The combined model merged the design of the image and EHR models by (i) passing chest radiographs through the frozen DenseNet-121 to extract image-based features, (ii) concatenating these features with EHR-based features, and (iii) passing these features through the output layer followed by a sigmoid output activation for each diagnosis.

Abbreviations: lock: frozen parameters; d: dimension of input EHR data; X: chest radiograph

**Table 1.** Accuracy of discharge diagnosis codes for identifying the etiology of acute respiratory failure.

| Diagnosis | Sensitivity | Specificity | Positive Predictive Value |
|---|---|---|---|
| **Pneumonia** | 0.78 | 0.77 | 0.61 |
| **Heart Failure** | 0.64 | 0.85 | 0.56 |
| **COPD** | 0.74 | 0.90 | 0.42 |

Accuracy of the discharge diagnosis codes is based on retrospective chart review.
Abbreviations: COPD: chronic obstructive pulmonary disease.

**Table 2:** Inter-annotator agreement in diagnosis among physicians.

| Agreement measure | Pneumonia | Heart Failure | COPD |
|---|---|---|---|
| **Cohen's Kappa** | 0.47 | 0.48 | 0.56 |
| **Raw Agreement** | 0.78 | 0.79 | 0.94 |

Abbreviations: COPD: chronic obstructive pulmonary disease.

**Figure 2.** Example chest radiograph heatmaps in patients where the model incorrectly diagnoses pneumonia, heart failure or COPD.

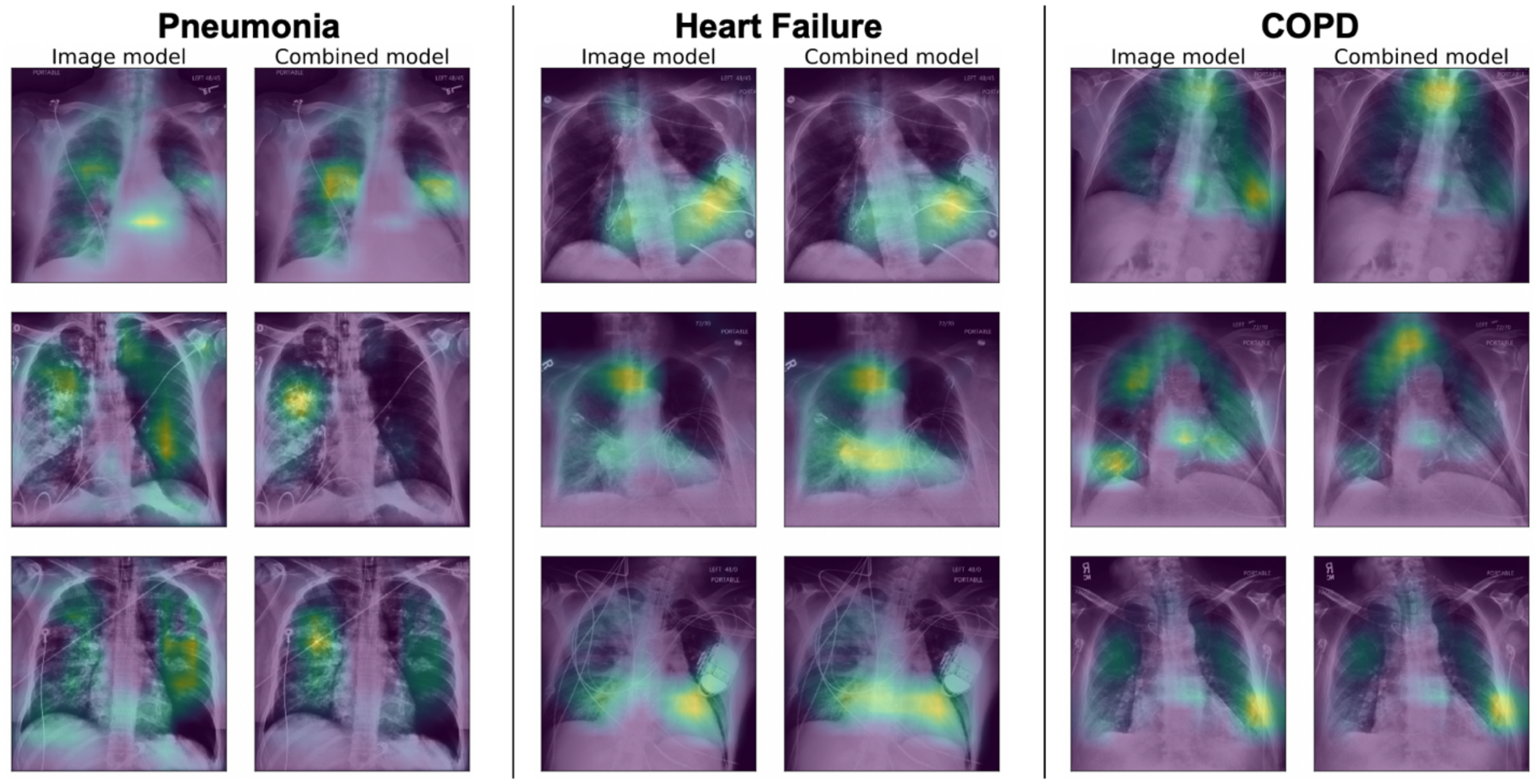

Chest radiographs are shown for patients that the model incorrectly classified as positive for each disease with high probability. The overlying heatmap generated by Grad-CAM highlights the regions that the model focuses on when estimating the likely diagnosis (blue: low contribution, yellow: high contribution). For both the image and combined models, the model looks at the lung regions for pneumonia and COPD and the heart region for heart failure. Heatmaps are normalized on individual images to highlight the most important areas of each image, therefore heatmap values should not be compared across images. Image processing was performed, including histogram equalization to increase contrast in the original images, and then images were resized to 512x512 pixels.
Abbreviations: COPD: chronic obstructive pulmonary disease.

**Table 3**. Performance of image, EHR and combined models on the internal held-out test set and external validation cohort in terms of AUPR.

| Cohort and Model | Pneumonia | Heart Failure | COPD | Macro-Average AUPR |
|---|---|---|---|---|
| **Internal chart review (n, % pos)** | **324 (322-324) 32% (32-36)** | **324 (322-324) 21% (20-23)** | **324 (322-324) 8% (5-8)** | **--** |
| Image | 0.58 (0.54-0.64) | 0.50 (0.44-0.57) | 0.43 (0.32-0.58) | 0.53 (0.46-0.57) |
| EHR | 0.60 (0.57-0.65) | 0.55 (0.44-0.60) | 0.39 (0.28-0.44) | 0.51 (0.48-0.53) |
| Combined | 0.67 (0.61-0.72) | 0.58 (0.53-0.64) | 0.59 (0.42-0.74) | 0.64 (0.55-0.67) |
| **Internal diagnosis codes + meds (n, % pos)** | **324 (322-324) 45% (37-46)** | **324 (322-324) 26% (22-29)** | **324 (322-324) 15% (15-16)** | **--** |
| Image | 0.62 (0.51-0.63) | 0.58 (0.56-0.59) | 0.36 (0.31-0.43) | 0.51 (0.48-0.53) |
| EHR | 0.60 (0.53-0.68) | 0.58 (0.51-0.63) | 0.37 (0.33-0.51) | 0.50 (0.49-0.57) |
| Combined | 0.65 (0.54-0.70) | 0.65 (0.63-0.66) | 0.46 (0.40-0.55) | 0.57 (0.56-0.62) |
| **External diagnosis codes + meds (n, % pos)** | **n=1774 18%** | **n=1774 11%** | **n=1774 4%** | **--** |
| Image | 0.28 (0.27-0.29) | 0.37 (0.35-0.39) | 0.14 (0.10-0.15) | 0.26 (0.25-0.27) |
| EHR | 0.25 (0.24-0.26) | 0.28 (0.21-0.31) | 0.15 (0.12-0.17) | 0.22 (0.21-0.25) |
| Combined | 0.28 (0.27-0.28) | 0.37 (0.35-0.40) | 0.19 (0.18-0.21) | 0.28 (0.27-0.29) |

Performance as determined based on the area under the precision-recall curve (AUPR). The internal cohort was randomly split five times into train (60%), validation (20%) and test (20%) sets. The median AUPR and AUPR range are reported for models trained on each split. The resulting five models were applied to the external cohort and the median AUPR and AUPR range are reported for models.
Abbreviations: COPD: chronic obstructive pulmonary disease.

**Table 4**. Performance of image, EHR and combined models on the internal train, validation, and test sets in terms of AUROC.

| Model and Data Split | Pneumonia | Heart Failure | COPD |
|---|---|---|---|
| **Combined** | | | |
| Train | 0.89 (0.88 - 0.92) | 0.90 (0.89 - 0.91) | 0.95 (0.94 - 0.97) |
| Validation | 0.79 (0.77 - 0.80) | 0.82 (0.80 - 0.85) | 0.89 (0.86 - 0.95) |
| Test | 0.79 (0.77 - 0.79) | 0.83 (0.77 - 0.84) | 0.88 (0.83 - 0.91) |
| **Image** | | | |
| Train | 0.77 (0.74 - 0.77) | 0.79 (0.78 - 0.82) | 0.85 (0.83 - 0.86) |
| Validation | 0.76 (0.72 - 0.78) | 0.78 (0.76 - 0.84) | 0.86 (0.79 - 0.90) |
| Test | 0.74 (0.72 - 0.75) | 0.80 (0.71 - 0.81) | 0.83 (0.76 - 0.90 |
| **EHR** | | | |
| Train | 0.88 (0.83 - 0.91) | 0.90 (0.85 - 0.93) | 0.92 (0.88 - 0.95) |
| Validation | 0.74 (0.73 - 0.77) | 0.77 (0.75 - 0.79) | 0.85 (0.83 - 0.91) |
| Test | 0.74 (0.70 - 0.76) | 0.79 (0.75 - 0.82) | 0.80 (0.76 - 0.84) |

Performance as determined based on the area under the receiver operator characteristic curve (AUROC) in the train, validation, and test sets. The cohort was randomly split five times into train (60%), validation (20%) and test (20%) sets. The median AUROC and AUROC range are reported for models trained on each split.
Abbreviations: COPD: chronic obstructive pulmonary disease.

**Figure 3**. Calibration plots and ECE (median and range) for the image, EHR, and combined models when applied to the internal MM cohort based on chart review.

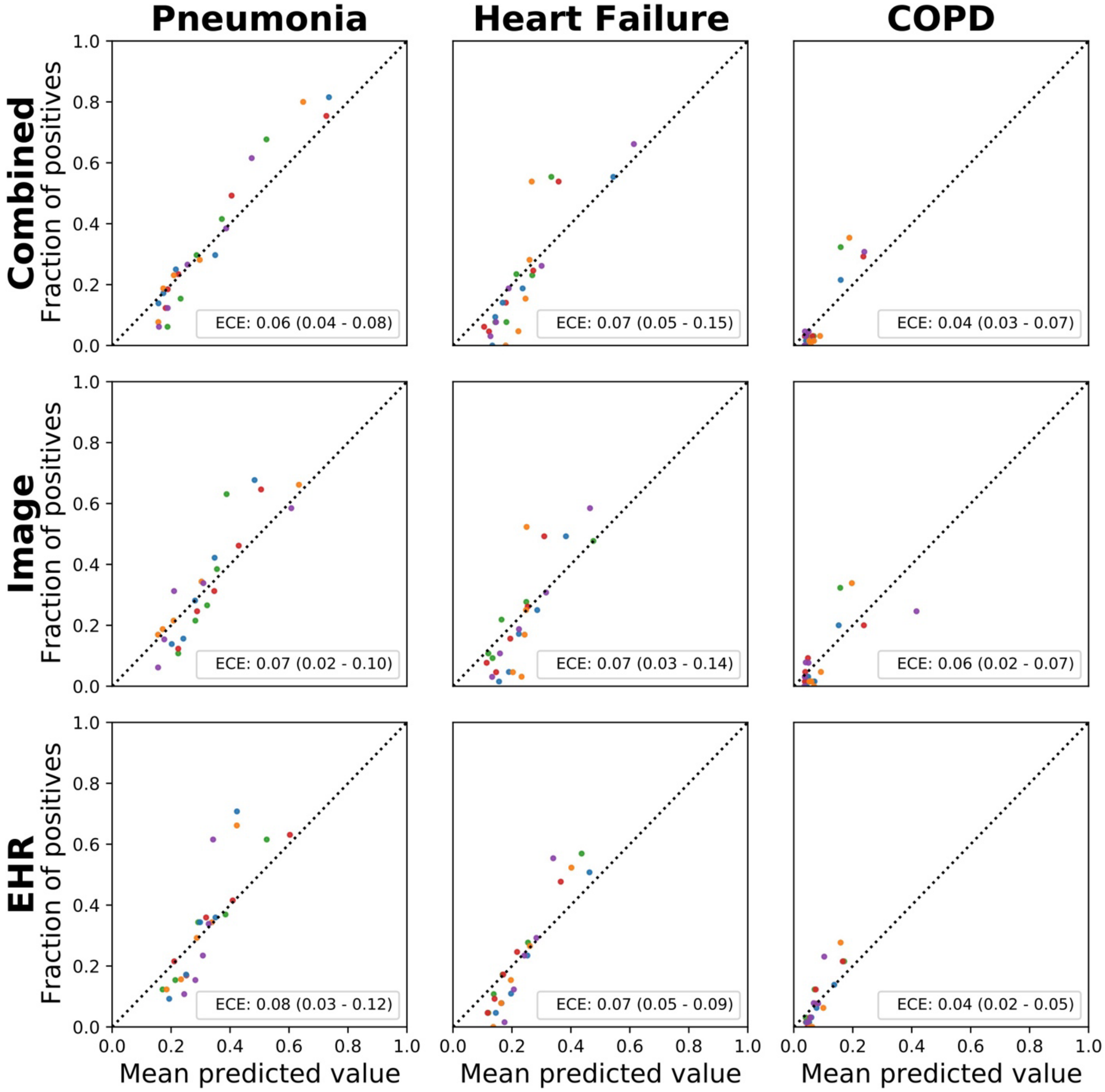

For each model and each disease, we divided the model predictions into quintiles and calculated the average prediction of each quintile. We then plotted the average prediction over the true probability of disease and computed a line of best fit to map the models' predictions to calibrated predictions. ECE was measured based on quintiles of predicted risk by calculating the average absolute difference between predicted risk and observed risk.
Abbreviations: COPD: chronic obstructive pulmonary disease; ECE: expected calibration error.

**Table 5**. Top 5 correlations measured using the Spearman's rank correlation coefficient between the presence of specific EHR variables and diagnoses

| Diagnosis | EHR Feature Name | Correlation |
|---|---|---|
| Pneumonia | Procalcitonin | 0.28 |
| | BNP | 0.19 |
| | Absolute lymphocyte count | 0.17 |
| | Monocytes | 0.17 |
| | Basophils | 0.17 |
| Heart Failure | BNP | 0.30 |
| | Tidal Volume Observed | -0.26 |
| | Tidal Volume Set | -0.26 |
| | Plateau Pressure | -0.25 |
| | Troponin-I | -0.23 |
| COPD | Tidal Volume Set | -0.19 |
| | BNP | 0.18 |
| | Tidal Volume Observed | -0.18 |
| | Plateau Pressure | -0.17 |
| | Respiratory Rate (Set) | -0.17 |

For each patient, each EHR feature out of the 68 total UM features was labelled as missing (0) or not (1). The correlation between missingness and each diagnosis was then measured. The presence of certain measures correlates with the prognostic importance of such measures in clinical practice. For example, the presence of procalcitonin correlated with a patient having pneumonia, and the presence of BNP correlated with a patient having heart failure. Although there aren't established EHR markers for COPD, we observe that the presence of tidal volume set has a negative correlation with COPD.
Abbreviations: COPD: chronic obstructive pulmonary disease.

**Table 6**. International Disease classification 10 codes for pneumonia, heart failure, and COPD.

| Diagnosis | ICD Code | Description |
|---|---|---|
| **Pneumonia** | J69.0 | Pneumonitis due to inhalation of food and vomit |
| | A48.1 | Legionnaires' disease |
| | J09.X1 | Influenza due to identified novel influenza A virus with pneumonia |
| | J10.00 | Influenza due to other identified influenza virus with unspecified type of pneumonia |
| | J10.01 | Influenza due to other identified influenza virus with the same other identified influenza virus pneumonia |
| | J10.08 | Influenza due to other identified influenza virus with other specified pneumonia |
| | J11.00 | Influenza due to unidentified influenza virus with unspecified type of pneumonia |
| | J11.08 | Influenza due to unidentified influenza virus with specified pneumonia |
| | J12.0 | Adenoviral pneumonia |
| | J12.1 | Respiratory syncytial virus pneumonia |
| | J12.2 | Parainfluenza virus pneumonia |
| | J12.3 | Human metapneumovirus pneumonia |
| | J12.81 | Pneumonia due to SARS-associated coronavirus |
| | J12.89 | Other viral pneumonia |
| | J12.9 | Viral pneumonia, unspecified |
| | J13 | Pneumonia due to Streptococcus pneumoniae |
| | J14 | Pneumonia due to Hemophilus influenzae |
| | J15.0 | Pneumonia due to Klebsiella pneumoniae |
| | J15.1 | Pneumonia due to Pseudomonas |
| | J15.20 | Pneumonia due to staphylococcus, unspecified |
| | J15.211 | Pneumonia due to Methicillin susceptible Staphylococcus aureus |
| | J15.212 | Pneumonia due to Methicillin resistant Staphylococcus aureus |
| | J15.29 | Pneumonia due to other staphylococcus |
| | J15.3 | Pneumonia due to streptococcus, group B |
| | J15.4 | Pneumonia due to other streptococci |
| | J15.5 | Pneumonia due to Escherichia coli |
| | J15.6 | Pneumonia due to other Gram-negative bacteria |
| | J15.7 | Pneumonia due to Mycoplasma pneumoniae |
| | J15.8 | Pneumonia due to other specified bacteria |
| | J15.9 | Unspecified bacterial pneumonia |
| | J16.0 | Chlamydial pneumonia |
| | J16.8 | Pneumonia due to other specified infectious organisms |
| | J18.0 | Bronchopneumonia, unspecified organism |
| | J18.1 | Lobar pneumonia, unspecified organism |
| | J18.8 | Other pneumonia, unspecified organism |
| | J18.9 | Pneumonia, unspecified organism |
| **Heart Failure** | I11.0 | Hypertensive heart disease with heart failure |
| | I13.0 | Hypertensive heart and chronic kidney disease with heart failure and stage 1 through stage 4 chronic kidney disease |
| | I13.2 | Hypertensive heart and chronic kidney disease with heart failure and with stage 5 chronic kidney disease, or end stage renal disease |
| | I50.1 | Left ventricular failure, unspecified |
| | I50.20 | Unspecified systolic (congestive) heart failure |
| | I50.21 | Acute systolic (congestive) heart failure |
| | I50.22 | Chronic systolic (congestive) heart failure |
| | I50.23 | Acute diastolic (congestive) heart failure |
| | I50.30 | Unspecified diastolic (congestive) heart failure |
| | I50.31 | Acute diastolic (congestive) heart failure |
| | I50.32 | Chronic diastolic (congestive) heart failure |
| | I50.33 | Acute on chronic diastolic (congestive) heart failure |
| | I50.40 | Unspecified combined systolic (congestive) and diastolic (congestive) heart failure |
| | I50.41 | Acute combined systolic (congestive) and diastolic (congestive) heart failure |
| | I50.42 | Chronic combined systolic (congestive) and diastolic (congestive) heart failure |
| | I50.43 | Acute on chronic combined systolic (congestive) and diastolic (congestive) heart failure |
| | I50.810 | Right heart failure, unspecified |
| | I50.811 | Acute right heart failure |
| | I50.812 | Chronic right heart failure |
| | I50.813 | Acute on chronic right heart failure |
| | I50.814 | Right heart failure due to left heart failure |
| | I50.82 | Biventricular heart failure |
| | I50.83 | High output heart failure |
| | I50.84 | End stage heart failure |
| | I50.89 | Other heart failure |
| | I50.9 | Heart failure, unspecified |
| **COPD** | J41.0 | Simple chronic bronchitis |
| | J41.1 | Mucopurulent chronic bronchitis |
| | J41.8 | Mixed simple and mucopurulent chronic bronchitis |
| | J42 | Unspecified chronic bronchitis |
| | J43.0 | Unilateral pulmonary emphysema [MacLeod's syndrome] |
| | J43.1 | Panlobular emphysema |
| | J43.2 | Centrilobular emphysema |
| | J43.8 | Other emphysema |
| | J43.9 | Emphysema, unspecified |
| | J44.0 | Chronic obstructive pulmonary disease with acute lower respiratory infection |
| | J44.1 | Chronic obstructive pulmonary disease with (acute) exacerbation |
| | J44.9 | Chronic obstructive pulmonary disease, unspecified |

Abbreviations: COPD: chronic obstructive pulmonary disease.

**Table 7**. Medications for pneumonia, heart failure, and COPD for the internal cohort.

| Diagnosis | Medication |
|---|---|
| Pneumonia | AMIKACIN <= 1000 MG IVPB |
| | AMOXICILLIN 200 MG-POTASSIUM CLAVULANATE 28.5 MG/5 ML ORAL SUSPENSION |
| | AMOXICILLIN 875 MG-POTASSIUM CLAVULANATE 125 MG TABLET |
| | AMPICILLIN-SULBACTAM 1.5 GRAM SOLUTION FOR INJECTION |
| | AMPICILLIN-SULBACTAM 3 GRAM SOLUTION FOR INJECTION |
| | AMPICILLIN-SULBACTAM ADD 3 G IN 100 ML NS |
| | AMPICILLIN-SULBACTAM IVPB (WITH ADAPTER) |
| | AMPICILLIN-SULBACTAM PRE ASSEMBLED 1.5 G IN 100 ML NS |
| | AZTREONAM INJECTION SYRINGE |
| | AZTREONAM IVPB |
| | AZTREONAM PRE ASSEMBLED 1 GM IN 100 ML NS |
| | AZTREONAM PRE ASSEMBLED 2 GM IN 100 ML NS |
| | CEFEPIME 1 GRAM SOLUTION FOR INJECTION |
| | CEFEPIME 1 GRAM/50 ML IN DEXTROSE (ISO-OSMOTIC) INTRAVENOUS PIGGYBACK |
| | CEFEPIME 2 GRAM SOLUTION FOR INJECTION |
| | CEFEPIME 2 GRAM/100 ML IN DEXTROSE (ISO-OSMOTIC) INTRAVENOUS PIGGYBACK |
| | CEFEPIME 2 GRAM/50 ML IN DEXTROSE 5 % INTRAVENOUS PIGGYBACK |
| | CEFEPIME ADD 1 G IN 50 ML D5W |
| | CEFEPIME ADD 2 G IN 50 ML D5W |
| | CEFEPIME ADD 2 G IN 50 ML D5W - EXTENDED INFUSION |
| | CEFEPIME IVPB |
| | CEFEPIME PRE ASSEMBLED 1 G IN 100 ML NS |
| | CEFEPIME PRE ASSEMBLED 2 G IN 100 ML NS |
| | CEFTAROLINE 400 MG IN NS 250 ML IVPB |
| | CEFTAROLINE 600 MG IN NS 250 ML IVPB |
| | CEFTAZIDIME 50 MG/0.5 ML SUBCONJUNCTIVAL |
| | CEFTAZIDIME INJECTION SYRINGE |
| | CEFTAZIDIME IVPB |
| | CEFTAZIDIME PRE ASSEMBLED 1 GM IN 100 ML NS |
| | CEFTAZIDIME PRE ASSEMBLED 2 GM IN 100 ML NS |
| | CEFTAZIDIME-AVIBACTAM IVPB |
| | CEFTOLOZANE-TAZOBACTAM IVPB |
| | CEFTRIAXONE 1 GRAM SOLUTION FOR INJECTION |
| | CEFTRIAXONE 1 GRAM/50 ML IN DEXTROSE (ISO-OSMOT) INTRAVENOUS PIGGYBACK |
| | CEFTRIAXONE 2 GRAM SOLUTION FOR INJECTION |
| | CEFTRIAXONE 2 GRAM/50 ML IN DEXTROSE (ISO-OSM) INTRAVENOUS PIGGYBACK |
| | CEFTRIAXONE ADD 1 G IN 50 ML D5W |
| | CEFTRIAXONE ADD 2 G IN 50 ML D5W |
| | CEFTRIAXONE INJECTION SYRINGE |
| | CEFTRIAXONE PRE ASSEMBLED 1 G IN 100 ML NS |
| | CEFTRIAXONE PRE ASSEMBLED 2 G IN 100 ML NS |
| | CEFUROXIME PRE ASSEMBLED 1.5 G IN 100 ML NS |
| | CIPROFLOXACIN 500 MG TABLET |
| | IMIPENEM-CILASTATIN (250 ML) |
| | IMIPENEM-CILASTATIN ADD 500 MG IN 100 ML NS |
| | LEVOFLOXACIN 250 MG TABLET |
| | LEVOFLOXACIN 250 MG/10 ML ORAL SOLUTION |
| | LEVOFLOXACIN 250 MG/50 ML IN 5 % DEXTROSE INTRAVENOUS PIGGYBACK |
| | LEVOFLOXACIN 500 MG TABLET |
| | LEVOFLOXACIN 500 MG/100 ML IN 5 % DEXTROSE INTRAVENOUS PIGGYBACK |
| | LEVOFLOXACIN 750 MG TABLET |
| | LEVOFLOXACIN 750 MG/150 ML IN 5 % DEXTROSE INTRAVENOUS PIGGYBACK |
| | MEROPENEM ADD 1 G IN 100 ML NS |
| | MEROPENEM INJECTION SYRINGE |
| | MEROPENEM IVPB |
| | MEROPENEM IVPB - EXTENDED INFUSION |
| | MEROPENEM PRE ASSEMBLED 500 MG IN 100 ML NS |
| | MOXIFLOXACIN 20 MG/ML ORAL SUSPENSION |
| | MOXIFLOXACIN 400 MG TABLET |
| | PIPERACILLIN-TAZOBACTAM 3.375 GRAM INTRAVENOUS SOLUTION |
| | PIPERACILLIN-TAZOBACTAM 3.375 GRAM/50 ML DEXTROSE(ISO-OS) IV PIGGYBACK |
| | PIPERACILLIN-TAZOBACTAM 4.5 GRAM INTRAVENOUS SOLUTION |
| | PIPERACILLIN-TAZOBACTAM 4.5 GRAM/100 ML DEXTROSE(ISO-OSM) IV PIGGYBACK |
| | PIPERACILLIN-TAZOBACTAM ADD 3.375 G IN 50 ML D5W |
| | PIPERACILLIN-TAZOBACTAM ADD 4.5 G IN 50 ML 0.9 % NS |
| | PIPERACILLIN-TAZOBACTAM ADD 4.5 G IN 50 ML D5W |
| | PIPERACILLIN-TAZOBACTAM ADD 4.5 G IN 50 ML D5W-EXTENDED INF |
| | PIPERACILLIN-TAZOBACTAM ADD 4.5 G IN D5W KIT |
| | PIPERACILLIN-TAZOBACTAM IVPB |
| | PIPERACILLIN-TAZOBACTAM PRE ASSEMBLED 2.25 GM IN 100 ML NS |
| | PIPERACILLIN-TAZOBACTAM PRE ASSEMBLED 3.375 GM IN 100 ML NS |
| | PIPERACILLIN-TAZOBACTAM PRE ASSEMBLED 4.5 GM IN 100 ML NS |
| | TIGECYCLINE IVPB |
| | TOBRAMYCIN </= 300 MG IVPB (WITH ADAPTER) |
| | TOBRAMYCIN <= 300 MG IN SODIUM CHLORIDE 0.9% IVPB |
| | TOBRAMYCIN > 300 MG IN SODIUM CHLORIDE 0.9% IVPB |
| | TOBRAMYCIN > 300 MG IVPB (WITH ADAPTER) |
| | VANCOMYCIN <= 1000 MG IVPB |
| | VANCOMYCIN <= 1000 MG IVPB - IN 0.9% NACL |
| | VANCOMYCIN > 1000 MG IVPB (250 ML NS) |
| | VANCOMYCIN 1 GM IVPB |
| | VANCOMYCIN 1 GRAM/200 ML IN 0.9 % SOD. CHLORIDE INTRAVENOUS PIGGYBACK |
| | VANCOMYCIN 1,000 MG INTRAVENOUS SOLUTION |
| | VANCOMYCIN 1.25 GRAM/150 ML IN 0.9 % SODIUM CHLORIDE INTRAVENOUS |
| | VANCOMYCIN 1.25 GRAM/250 ML IN DEXTROSE 5 % INTRAVENOUS |
| | VANCOMYCIN 1.5 GRAM/150 ML IN 0.9 % SODIUM CHLORIDE INTRAVENOUS |
| | VANCOMYCIN 500 MG IV SYRINGE NS (OR) |
| | VANCOMYCIN 500 MG IV SYRINGE OR |
| | VANCOMYCIN 750 MG/150 ML IN 0.9 % SODIUM CHLORIDE INTRAVENOUS |
| | VANCOMYCIN ADD 1 G IN 100 ML D5W |
| | VANCOMYCIN ADD 1 G IN 100 ML NS |
| | VANCOMYCIN ADD 500 MG IN 100 ML NS |
| | VANCOMYCIN INJECTION 1000 MG VIAL (OR) |
| | VANCOMYCIN PRE ASSEMBLED 750 MG IN 100 ML NS |
| **Heart Failure** | BUMETANIDE 0.25 MG/ML INJECTION SOLUTION |
| | BUMETANIDE 0.5 MG TABLET |
| | BUMETANIDE 1 MG TABLET |
| | BUMETANIDE 2 MG TABLET |
| | BUMETANIDE INFUSION ADULT |
| | BUMETANIDE INFUSION CVICU |
| | BUMETANIDE IVPB (ADULT) |
| | CHLOROTHIAZIDE 250 MG TABLET |
| | CHLOROTHIAZIDE 250 MG/5 ML ORAL SUSPENSION |
| | CHLOROTHIAZIDE INJECTION |
| | CHLOROTHIAZIDE IVPB ADULT |
| | CHLORTHALIDONE 25 MG TABLET |
| | ETHACRYNATE INJECTION |
| | FUROSEMIDE 10 MG/ML INJECTION SOLUTION |
| | FUROSEMIDE 10 MG/ML ORAL LIQ (WRAPPER) |
| | FUROSEMIDE 20 MG TABLET |
| | FUROSEMIDE 40 MG TABLET |
| | FUROSEMIDE 80 MG TABLET<br>FUROSEMIDE INFUSION<br>FUROSEMIDE INFUSION - PEDS - NO DILUENT<br>FUROSEMIDE INFUSION SYRINGE<br>FUROSEMIDE INJECTION SYRINGE (PEDS)<br>FUROSEMIDE IVPB (ADULT)<br>TORSEMIDE 100 MG TABLET<br>TORSEMIDE 20 MG TABLET<br>TORSEMIDE 50 MG (1/2 100 MG) TABLET |
| **COPD** | DEXAMETHASONE 0.5 MG/5 ML ORAL SOLUTION<br>DEXAMETHASONE 1 MG TABLET<br>DEXAMETHASONE 4 MG TABLET<br>DEXAMETHASONE 4 MG/ML INJECTION SOLUTION<br>DEXAMETHASONE IVPB<br>HYDROCORTISONE 10 MG TABLET<br>HYDROCORTISONE 100 MG SOLUTION FOR INJECTION<br>HYDROCORTISONE 20 MG TABLET<br>HYDROCORTISONE 5 MG TABLET<br>HYDROCORTISONE INFUSION ADULT<br>HYDROCORTISONE IVPB<br>HYDROCORTISONE SOD SUCCINATE (PF) 100 MG/2 ML SOLUTION FOR INJECTION<br>HYDROCORTISONE SOD SUCCINATE (SOLU-CORTEF) INJECTION 100 G VIALH (OR)<br>METHYLPREDNISOLONE 16 MG TABLET<br>METHYLPREDNISOLONE 4 MG TABLET<br>METHYLPREDNISOLONE INJECTION SYRINGE<br>METHYLPREDNISOLONE IVPB (100 ML) IN D5W<br>METHYLPREDNISOLONE IVPB (50 ML) IN D5W<br>METHYLPREDNISOLONE SOD SUCC (PF) 125 MG/2 ML SOLUTION FOR INJECTION<br>METHYLPREDNISOLONE SOD SUCC ADD 1,000 MG IN 100 ML D5W<br>METHYLPREDNISOLONE SOD SUCC PRE ASSEMBLED 500 MG IN 100 ML NS<br>METHYLPREDNISOLONE SODIUM SUCC 125 MG SOLUTION FOR INJECTION<br>PREDNISOLONE 5 MG TABLET<br>PREDNISOLONE SODIUM PHOSPHATE 15 MG/5 ML ORAL SOLUTION<br>PREDNISONE 1 MG TABLET<br>PREDNISONE 10 MG TABLET<br>PREDNISONE 2.5 MG TABLET<br>PREDNISONE 20 MG TABLET<br>PREDNISONE 5 MG TABLET |
| | PREDNISONE 5 MG/5 ML ORAL SOLUTION<br>PREDNISONE 5 MG/ML ORAL SUSPENSION |

**Table 8**. Medications for pneumonia, heart failure, and COPD for the external cohort.

| Diagnosis | Medication |
|---|---|
| Pneumonia | Amoxicillin-Clavulanate Susp. |
| | Amoxicillin-Clavulanic Acid |
| | Ampicillin-Sulbactam |
| | Azithromycin |
| | Aztreonam |
| | CIPROFLOXACIN |
| | CefTAZidime |
| | CefTRIAXone |
| | CefePIME |
| | Cefpodoxime Proxetil |
| | Ceftaroline |
| | CeftriaXONE |
| | Ciprofloxacin |
| | Clindamycin |
| | DiCLOXacillin |
| | Doxycycline Hyclate |
| | Ertapenem Sodium |
| | Imipenem-Cilastatin |
| | Levofloxacin |
| | Linezolid |
| | Meropenem |
| | Moxifloxacin |
| | Piperacillin-Tazobactam |
| | Tigecycline |
| | Tobramycin |
| | Vancomycin |
| | ceftazidime-avibactam |
| | gatifloxacin |
| | imipenem-cilastatin |
| | moxifloxacin |
| Heart Failure | Bumetanide |
| | Chlorothiazide |
| | Ethacrynate Sodium |
| | Furosemid |
| | Furosemide |
| | Furosemide in 0.9% Sodium Chloride |
| | Furosemide-Heart Failure<br>Metolazone<br>Torsemide |
| **COPD** | MethylPREDNISolone Sodium Succ<br>Methylprednisolone<br>PredniSONE<br>predniSONE |

Abbreviations: COPD: chronic obstructive pulmonary disease.

**Table 9.** Feature mapping from Michigan Medicine to Beth Israel Deaconess

| | MM Feature Name | MIMIC-IV/BIDMC Feature Name |
|---|---|---|
| Vital Signs | Diastolic blood pressure | Arterial Blood Pressure diastolic |
| | Diastolic blood pressure | ART BP Diastolic |
| | Diastolic blood pressure | Non Invasive Blood Pressure diastolic |
| | Diastolic blood pressure | Manual Blood Pressure Diastolic Left |
| | Fraction inspired oxygen | Inspired O2 Fraction |
| | Heart Rate | Heart Rate |
| | Height | Height (cm) |
| | Mean blood pressure | Arterial Blood Pressure mean |
| | Mean blood pressure | ART BP mean |
| | Mean blood pressure | IABP Mean |
| | Mean blood pressure | Non Invasive Blood Pressure mean |
| | Pulse oximetry | O2 saturation pulse oxymetry |
| | Peak inspiratory pressure | Peak Insp. Pressure |
| | Positive end-expiratory pressure Set | PEEP set |
| | Respiratory rate | Respiratory Rate |
| | Respiratory rate | Respiratory Rate (spontaneous) |
| | Respiratory rate | Spont RR |
| | Respiratory rate | Respiratory Rate (Total) |
| | Respiratory Rate (Set) | Respiratory Rate (Set) |
| | Systolic blood pressure | Arterial Blood Pressure systolic |
| | Systolic blood pressure | ART BP Systolic |
| | Systolic blood pressure | Non Invasive Blood Pressure systolic |
| | Systolic blood pressure | Manual Blood Pressure Systolic Left |
| | Systolic blood pressure | Manual Blood Pressure Systolic Right |
| | Temperature (C) | Temperature Celsius |
| | Weight | Admission Weight (Kg) |
| | Plateau Pressure | Plateau Pressure |
| | Tidal Volume Observed | Tidal Volume (observed) |
| | Tidal Volume Set | Tidal Volume (set) |
| | Tidal Volume Spontaneous | Tidal Volume (spontaneous) |
| Lab Results | Alanine aminotransferase | ALANINE AMINOTRANSFERASE (ALT) |
| | Albumin | ALBUMIN |
| | Alkaline phosphate | ALKALINE PHOSPHATASE |
| | Asparate aminotransferase | ASPARATE AMINOTRANSFERASE (AST) |
| | Basophils | BASOPHILS % |
| | Bicarbonate | BICARBONATE |
| | Bilirubin (conjugated) | BILIRUBIN, DIRECT |
| | Bilirubin (total) | BILIRUBIN, TOTAL |
| | Bilirubin (unconjugated) | BILIRUBIN, INDIRECT |
| | Blood urea nitrogen | UREA NITROGEN |
| | Calcium (total) | CALCIUM, TOTAL |
| | Calcium ionized | FREE CALCIUM |
| | Chloride | CHLORIDE |
| | Cholesterol (total) | CHOLESTEROL, TOTAL |
| | Cholesterol (HDL) | CHOLESTEROL, HDL |
| | Creatinine | CREATININE |
| | Eosinophils (blood) | EOSINOPHILS % |
| | Glucose | GLUCOSE |
| | Hematocrit | HEMATOCRIT |
| | Hemoglobin | HEMOGLOBIN |
| Lab Results | Lactate | LACTATE |
| | Lactate dehydrogenase | LACTATE DEHYDROGENASE (LD) |
| | Lymphocytes | LYMPHOCYTES |
| | Lymphocytes (absolute) | ABSOLUTE LYMPHOCYTE COUNT |
| | Magnesium | MAGNESIUM |
| | Mean corpuscular hemoglobin | MCH |
| | Mean corpuscular hemoglobin concentration | MCHC |
| | Mean corpuscular volume | MCV |
| | Monocytes | MONOCYTES |
| | Neutrophils | NEUTROPHILS |
| | partial pressure of oxygen | PO2 |
| | Partial pressure of carbon dioxide | PCO2 |
| | Oxygen saturation | OXYGEN SATURATION |
| | Partial thromboplastin time | PTT |
| | pH | PH |
| | Phosphate | PHOSPHATE |
| | Platelet Count | PLATELET COUNT |
| | Potassium | POTASSIUM |
| | Prothrombin time | INR(PT) |
| | Red blood cell count | RED BLOOD CELLS |
| | Sodium | SODIUM |
| | Troponin-I | TROPONIN T |
| | White blood cell count | WHITE BLOOD CELLS |
| | Fibrinogen | FIBRINOGEN, FUNCTIONAL |
| | BNP | NTproBNP |
| | Procalcitonin | --- |
| Static Variables | Age | anchor_age |
| | Gender | gender |
| | Race | ethnicity |

# TRIPOD Checklist: Prediction Model Development and Validation

| **Section/Topic** | | | **Checklist Item** | **Page** |
|---|---|---|---|---|
| **Title and abstract** | | | | |
| Title | 1 | D;V | Identify the study as developing and/or validating a multivariable prediction model, the target population, and the outcome to be predicted. | 1 |
| Abstract | 2 | D;V | Provide a summary of objectives, study design, setting, participants, sample size, predictors, outcome, statistical analysis, results, and conclusions. | 2-3 |
| **Introduction** | | | | |
| Background and objectives | 3a | D;V | Explain the medical context (including whether diagnostic or prognostic) and rationale for developing or validating the multivariable prediction model, including references to existing models. | 4 |
| | 3b | D;V | Specify the objectives, including whether the study describes the development or validation of the model or both. | 4-5 |
| **Methods** | | | | |
| Source of data | 4a | D;V | Describe the study design or source of data (e.g., randomized trial, cohort, or registry data), separately for the development and validation data sets, if applicable. | 5-6 |
| | 4b | D;V | Specify the key study dates, including start of accrual; end of accrual; and, if applicable, end of follow-up. | 5 |
| Participants | 5a | D;V | Specify key elements of the study setting (e.g., primary care, secondary care, general population) including number and location of centres. | 5 |
| | 5b | D;V | Describe eligibility criteria for participants. | 5, Supplement 2 |
| | 5c | D;V | Give details of treatments received, if relevant. | NA |
| Outcome | 6a | D;V | Clearly define the outcome that is predicted by the prediction model, including how and when assessed. | 6,8 |
| | 6b | D;V | Report any actions to blind assessment of the outcome to be predicted. | Supplement 2 |
| Predictors | 7a | D;V | Clearly define all predictors used in developing or validating the multivariable prediction model, including how and when they were measured. | 7, Supplement 2 |
| | 7b | D;V | Report any actions to blind assessment of predictors for the outcome and other predictors. | Supplement 2 |
| Sample size | 8 | D;V | Explain how the study size was arrived at. | Supplement 2 |
| Missing data | 9 | D;V | Describe how missing data were handled (e.g., complete-case analysis, single imputation, multiple imputation) with details of any imputation method. | 7, Supplement 2 |
| Statistical analysis methods | 10a | D | Describe how predictors were handled in the analyses. | 8-9. Supplement 2-3 |
| | 10b | D | Specify type of model, all model-building procedures (including any predictor selection), and method for internal validation. | 8-10, Supplement 2-4 |
| | 10c | V | For validation, describe how the predictions were calculated. | 8-10 |
| | 10d | D;V | Specify all measures used to assess model performance and, if relevant, to compare multiple models. | 9-10 |
| | 10e | V | Describe any model updating (e.g., recalibration) arising from the validation, if done. | 9, Supplement 3 |
| Risk groups | 11 | D;V | Provide details on how risk groups were created, if done. | NA |
| Development vs. validation | 12 | V | For validation, identify any differences from the development data in setting, eligibility criteria, outcome, and predictors. | 6 |
| **Results** | | | | |
| Participants | 13a | D;V | Describe the flow of participants through the study, including the number of participants with and without the outcome and, if applicable, a summary of the follow-up time. A diagram may be helpful. | 11-12 |
| | 13b | D;V | Describe the characteristics of the participants (basic demographics, clinical features, available predictors), including the number of participants with missing data for predictors and outcome. | 11-12 |
| | 13c | V | For validation, show a comparison with the development data of the distribution of important variables (demographics, predictors and outcome). | 11-12 |
| Model development | 14a | D | Specify the number of participants and outcome events in each analysis. | 11-12 |
| | 14b | D | If done, report the unadjusted association between each candidate predictor and outcome. | NA |
| Model specification | 15a | D | Present the full prediction model to allow predictions for individuals (i.e., all regression coefficients, and model intercept or baseline survival at a given time point). | 21, 23 |
| | 15b | D | Explain how to the use the prediction model. | 19-20 |
| Model performance | 16 | D;V | Report performance measures (with CIs) for the prediction model. | 14-17, Supplement 5-10 |
| Model-updating | 17 | V | If done, report the results from any model updating (i.e., model specification, model performance). | Supplement 9 |
| **Discussion** | | | | |
| Limitations | 18 | D;V | Discuss any limitations of the study (such as nonrepresentative sample, few events per predictor, missing data). | 21-22 |
| Interpretation | 19a | V | For validation, discuss the results with reference to performance in the development data, and any other validation data. | 16-17, 22 |
| | 19b | D;V | Give an overall interpretation of the results, considering objectives, limitations, results from similar studies, and other relevant evidence. | 18-20 |
| Implications | 20 | D;V | Discuss the potential clinical use of the model and implications for future research. | 20-22 |
| **Other information** | | | | |
| Supplementary information | 21 | D;V | Provide information about the availability of supplementary resources, such as study protocol, Web calculator, and data sets. | 21, 23 |
| Funding | 22 | D;V | Give the source of funding and the role of the funders for the present study. | 22 |

*Items relevant only to the development of a prediction model are denoted by D, items relating solely to a validation of a prediction model are denoted by V, and items relating to both are denoted D;V. We recommend using the TRIPOD Checklist in conjunction with the TRIPOD Explanation and Elaboration document.

## Supplement References